\documentclass[conference]{IEEEtran}
\IEEEoverridecommandlockouts
\usepackage{cite}
\usepackage{amsmath,amssymb,amsfonts}
\usepackage{algorithmic}
\usepackage{graphicx}
\usepackage{xcolor} 
\definecolor{lightblue}{RGB}{30,144,255}
\usepackage[hidelinks]{hyperref}
\hypersetup{
    colorlinks=true,
    linkcolor=lightblue,
    citecolor=lightblue,
    urlcolor=lightblue
}
\usepackage{textcomp}
\usepackage{booktabs}
\usepackage{colortbl}
\usepackage[table]{xcolor}
\usepackage{xcolor}
\usepackage{float}
\usepackage{caption}
\usepackage[hidelinks]{hyperref}
\usepackage{doi}
\usepackage{array}
\usepackage{fancyhdr}
\usepackage{enumitem}
\def\BibTeX{{\rm B\kern-.05em{\sc i\kern-.025em b}\kern-.08em
    T\kern-.1667em\lower.7ex\hbox{E}\kern-.125emX}}

\begin{document}

% \title{Integrating Deep Forecasting and Reinforcement Learning for Wildfire Prediction and Containment}
% \title{Learning to Predict and Contain Wildfires: A Reinforcement Learning and Simulation-Based Approach}
\title{Spatiotemporal Wildfire Prediction and Reinforcement Learning for Helitack Suppression}

%A Deep Learning and Reinforcement Framework for Wildfire Forecasting and Tactical Suppression

%An End-to-End System for Wildfire Prediction and RL-Guided Suppression

%Spatiotemporal Forecasting and Reinforcement Learning for Wildfire Helitack Control

\author{
\IEEEauthorblockN{Shaurya Mathur, Shreyas Bellary Manjunath, Nitin Kulkarni, and Alina Vereshchaka}
\IEEEauthorblockA{Department of Computer Science and Engineering \\
University at Buffalo \\
Buffalo, New York, USA}
\IEEEauthorblockA{\{smathur4, sbellary, nitinvis, avereshc\}@buffalo.edu}
}

\maketitle
\thispagestyle{fancy}
\fancyhf{} % Clear header/footer
\renewcommand{\headrulewidth}{0pt} % Remove header rule
\fancyfoot[c]{%
  \parbox{\textwidth}{%
  \centering \scriptsize
  \copyright~2025 IEEE. Personal use of this material is permitted. Permission from IEEE must be obtained for all other uses, in any current or future media, including reprinting/republishing this material for advertising or promotional purposes, creating new collective works, for resale or redistribution to servers or lists, or reuse of any copyrighted component of this work in other works.
  }
}

% To do - Alina - Finalize fig 1 , reduce number of bullets, Finalize fig 4, Combine part 6 & 7 
% To do - us - add caption for ppo fig, fig 5 changes and caption, update web interface part where we talk general  part

\begin{figure*}[!t]
\centerline{\includegraphics[width=1\textwidth]{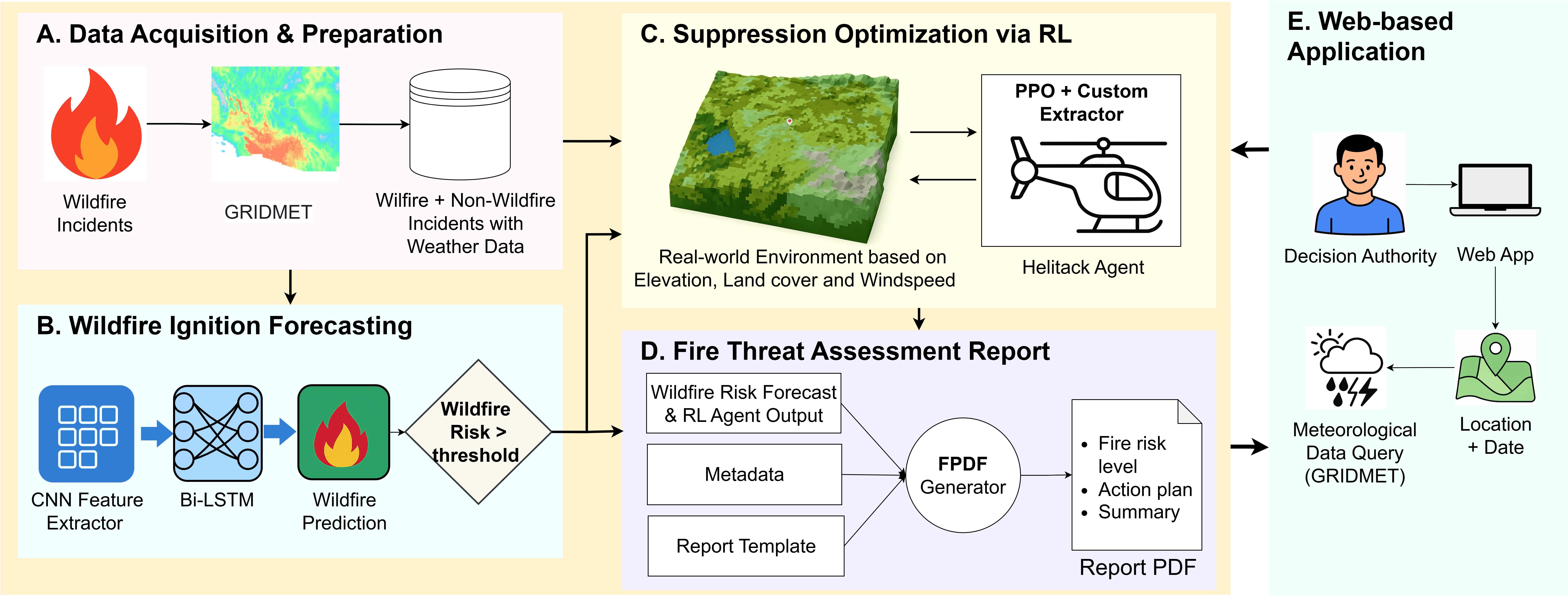}}
\caption{\textit{FireCastRL} framework is a wildfire forecasting and mitigation system containing five stages: (A) acquisition of IRWIN database and GRIDMET weather data, (B) prediction of wildfire ignition using CNN-LSTM architecture, (C) simulation of RL-based helitack suppression based on real-world terrain and environmental data, (D) generation of fire threat assessment report with risk forecasts and strategic recommendations, and (E) web-based application.}
% \caption{Inferno Tactics: Stage 1 predicts ignition hotspots; Stage 2 embeds each hotspot in a 3-D, physics-aware simulation where an RL Helitack agent tests suppression tactics; Stage 3 converts the roll-out into a stakeholder-ready brief.}
% \vspace{-0.35 cm}
\label{fig:pipeline}
\end{figure*}

\begin{abstract}
Wildfires are growing in frequency and intensity, devastating ecosystems and communities while causing billions of dollars in suppression costs and economic damage annually in the U.S. Traditional wildfire management is mostly reactive, addressing
fires only after they are detected. We introduce \textit{FireCastRL}, a proactive artificial intelligence (AI) framework that combines wildfire forecasting with intelligent suppression strategies.
Our framework first uses a deep spatiotemporal model to predict wildfire ignition. For high-risk predictions, we deploy a pre-trained reinforcement learning (RL) agent to execute real-time suppression tactics with helitack units inside a physics-informed 3D simulation. The framework generates a threat assessment report to help emergency responders optimize resource allocation and planning. In addition, we are publicly releasing a large-scale, spatiotemporal dataset containing $\mathbf{9.5}$ million samples of environmental variables for wildfire prediction. Our work demonstrates how deep learning and RL can be combined to support both forecasting and tactical wildfire response. More details can be found at \href{https://sites.google.com/view/firecastrl}{https://sites.google.com/view/firecastrl}.

\end{abstract}

% \begin{IEEEkeywords} Wildfire Prediction, Reinforcement Learning, Spatiotemporal Modeling, Physics-Informed Simulation, Time-Series Analysis, Disaster Response \end{IEEEkeywords}


\section{Introduction}

Wildfires in the U.S. are growing more frequent, intense, and costly. In $2023$ alone, they caused an estimated $\$14.7$ billion in direct property losses and required over $\$3$ billion in suppression spending~\cite{nifc2023suppression, iii2024property_loss}. When indirect damages and long-term effects are included, the total annual economic burden has been estimated at up to $\$893$ billion~\cite{jec2023wildfire}. This highlights the urgent need for more advanced and efficient wildfire management strategies to mitigate future losses.

Traditional wildfire detection methods (e.g., watchtowers, ground sensors, and satellite scans) are inherently reactive. They trigger alarms only after fires are visible, often leaving little time for effective containment, especially in remote or rugged areas. To mitigate risk more effectively, we need systems that forecast where and when fires are likely to start, and simulate how they might spread.

Recent advances in Earth observation data, deep learning, and simulation tools facilitate the development of novel approaches for predictive wildfire management. Deep learning models can learn patterns in environmental variables to forecast fire ignition days in advance, while reinforcement learning (RL) agents can learn optimal suppression strategies. 

% Studies~\cite{jonnalagadda2024segnet} highlight how AI and ML are improving wildfire management by improving forecasting, early detection, and fire spread modeling through the integration of datasets from satellites, sensors, and weather systems.

Ongoing efforts aim to integrate deep learning into wildfire monitoring and response. Google’s FireSat\cite{googlefiresat2023} project uses satellite imagery and deep learning models to detect active wildfires in real time, issuing alerts to first responders and governments globally. Similarly, Canada’s WildfireSat program and tools like National FireGuard focus on spaceborne detection and situational awareness.
While these systems excel at early detection, their primary scope is situational awareness rather than the downstream tasks of strategic response planning or autonomous suppression. In this work, we present \textit{FireCastRL}, our proposed framework that focuses on integrating wildfire prediction using deep learning methods and mitigation steps using RL, combining them into a single system. Our contributions include:

\begin{itemize}
    \item Publicly released, large-scale, spatiotemporal dataset containing $9.5$ million samples of environmental variables for wildfire prediction derived from GRIDMET~\cite{gridmet} and IRWIN. This dataset supports research on wildfire forecasting and AI-driven disaster response.
    \item A deep spatiotemporal forecasting model for predicting wildfire ignition based on patterns in historical environmental data.
    \item A physics-informed wildfire simulation environment that uses real land cover, elevation, and mesoscale wind fields. The engine combines a browser-based fire simulator for RL agent interaction.
    \item An RL-based helitack control policy trained with Proximal Policy Optimization (PPO). The agent learns to deploy aerial suppression (helitack) in dynamic wildfire environments.
    \item An end-to-end pipeline that generates a fire threat assessment report for emergency responders and decision-makers containing predicted ignition coordinates, burn trajectory, suppression sequence, and response recommendations.
\end{itemize}

\section{Related Work}

\subsubsection{Wildfire Prediction}
Recent models utilize MODIS satellite data and topographic variables with convolutional neural networks (CNNs) and recurrent neural networks to improve spatial and temporal generalization \cite{gollner2020deep, zhang2022wildfire}. FireCast~\cite{radke2019firecast}, a deep learning system, predicts near-future wildfire spread. While these works focus on predicting the progression of an already active fire without downstream integration into decision-making systems, our work predicts where a fire is likely to start, and subsequently uses RL to simulate and optimize an effective suppression response.

\begin{table*}[t]
\centering
\caption{Snapshot of the dataset showing location, date, wildfire labels, and environmental variables, including meteorological measurements.}
% \tiny % Makes the font very small to fit the wide table
\setlength{\tabcolsep}{3 pt}
\renewcommand{\arraystretch}{1.2}
  % \resizebox{\columnwidth}{!}{}
\begin{tabular}{|c|c|c|c|c|c|c|c|c|c|c|c|c|c|c|c|c|c|c|}
\hline
\textbf{latitude} & \textbf{longitude} & \textbf{datetime} & \textbf{Wildfire} & \textbf{pr} & \textbf{rmax} & \textbf{rmin} & \textbf{sph} & \textbf{srad} & \textbf{tmmn} & \textbf{tmmx} & \textbf{vs} & \textbf{bi} & \textbf{fm100} & \textbf{fm1000} & \textbf{erc} & \textbf{etr} & \textbf{pet} & \textbf{vpd} \\
\hline

48.128431 & -97.276685 & 2018-08-15 & No & 0.0 & 78.6 & 14.9 & 0.00582 & 272.6 & 282.0 & 301.6 & 3.0 & 40.0 & 10.2 & 12.2 & 54.0 & 7.5 & 5.5 & 1.59 \\
\hline

48.128431 & -97.276685 & 2018-08-16 & No & 0.0 & 80.4 & 13.9 & 0.00676 & 264.0 & 283.9 & 304.9 & 3.0 & 40.0 & 9.7 & 12.0 & 56.0 & 8.2 & 5.9 & 1.93 \\
\hline

% 48.128431 & -97.276685 & 2018-08-17 & No & 0.0 & 70.9 & 20.4 & 0.00672 & 265.6 & 285.8 & 300.7 & 3.1 & 40.0 & 9.2 & 11.9 & 56.0 & 7.2 & 5.3 & 1.51 \\
% \hline

\multicolumn{19}{|c|}{...} \\
\hline

% 37.920118 & -120.413184 & 2017-02-02 & No & 6.4 & 93.1 & 67.0 & 0.00764 & 35.7 & 282.5 & 289.0 & 8.7 & 0.0 & 16.0 & 25.1 & 8.0 & 2.5 & 1.6 & 0.29 \\
% \hline

% 37.920118 & -120.413184 & 2017-02-03 & No & 11.9 & 87.9 & 62.0 & 0.00767 & 73.7 & 283.4 & 290.2 & 8.0 & 0.0 & 19.0 & 25.9 & 5.0 & 3.2 & 2.1 & 0.39 \\
% \hline

37.920118 & -120.413184 & 2017-02-04 & No & 0.0 & 99.5 & 59.8 & 0.00713 & 85.8 & 280.5 & 289.6 & 2.8 & 19.0 & 17.2 & 25.4 & 14.0 & 1.9 & 1.4 & 0.33 \\
\hline

\end{tabular}
\label{tab:dataset_snapshot}
\end{table*}

\subsubsection{Fire Spread Simulation}
The Rothermel model \cite{rothermel1972} approximates the rate of fire spread based on fuel, wind, and elevation. Standard forest fire simulators like FARSITE~\cite{finney2004farsite} use this model to project wildfire behavior over real terrains. While accurate, these tools are computationally intensive and rarely used in closed-loop decision-making pipelines. Our approach draws inspiration from these simulators and integrates them into a real-time, RL-compatible environment.

\subsubsection{RL for Disaster Mitigation}
RL has been applied to a range of decision-making problems in uncertain environments, including emergency response, urban evacuation \cite{li2021reinforcement}, natural disasters \cite{vereshchaka2019reducing}, and epidemiological response optimization \cite{kulkarni2022optimizing}. Research in wildfire suppression using RL is limited but growing. For example, Julian et al. \cite{julian2020wildfire} modeled fire containment with discrete RL in grid environments. However, few studies incorporate realistic terrain or dynamics, and even fewer combine forecasting and mitigation in a single system.

\section{Proposed Framework}
We introduce \textit{FireCastRL} framework (Fig.~\ref{fig:pipeline}), an end-to-end system that connects wildfire prediction with strategic response planning. First, a deep learning model forecasts ignition risk based on a large-scale spatiotemporal dataset. High-risk predictions launch an RL simulation where a pre-trained agent deploys helitack suppression strategies in a physics-informed 3D environment. The prediction and simulation results are then compiled into a fire threat assessment report for emergency responders.

\subsection{Data Acquisition and Preparation}
We create a spatiotemporal wildfire dataset by combining high-resolution incident data from Integrated Reporting of Wildfire Information (IRWIN) with meteorological sequences from GRIDMET~\cite{gridmet}. This involves incident deduplication, negative sample synthesis, environmental feature extraction, and geographic-temporal validation.
\vspace{0.1em}

\subsubsection{Wildfire Incident Collection and Filtering}

Our dataset is based on the IRWIN database~\cite{IRWIN_database}, which catalogs all reported wildfire events across the U.S. from January 2014 to April 2025. The raw dataset consists of $348\,604$ wildfire reports, each tagged with metadata such as discovery timestamp and geographic coordinates.

A key challenge with this raw data is that wildfire events often appear multiple times in the database due to phased updates, secondary flare-ups, or redundant reports, thus introducing noise that can bias AI models. To isolate unique ignition events from data points generated by ongoing fire spread, we applied a multi-stage filtering procedure:

\begin{enumerate}
    \item For any given date, we retain only ignition events \(\geq~5~\mathrm{km}\) apart, using a sliding acceptance window that scans incidents sequentially and discards nearby duplicates.
    \item We enforce a minimum gap of $2$ hours between retained ignitions to remove short-term updates of the same fire.
    \item All retained coordinates are verified to lie within the Continental United States (CONUS) bounding box \([24.4^\circ\mathrm{N}, 49.4^\circ\mathrm{N}] \times [-125.0^\circ\mathrm{W}, -66.9^\circ\mathrm{W}]\) to cross-check inclusion within the U.S. boundaries.
\end{enumerate}

This process yields a spatiotemporally independent dataset of $50\,720$ positive wildfire ignition events.

\subsubsection{Dataset Balancing}
A challenge is that the dataset only provides positive labels (wildfire ignitions), with no corresponding data for where wildfires did not occur. To address this inherent class imbalance, we introduce a three-tier negative sampling framework to generate realistic non-wildfire samples:

\begin{itemize}
    \item Far Negatives (spatial isolation): $5000$ U.S.\ coordinates \(\geq~100~\mathrm{km}\) from a wildfire event, sampled without temporal constraints, representing low-risk historical zones.
    \item Near Negatives (temporal displacement): $35\,000$ samples within a $100$ $\mathrm{km}$ radius of a positive ignition, but offset by $90-150$ days to represent seasonal variability.
    \item Yearly Negatives (historical context): $36\,000$ samples at the same coordinates as positive events, offset by a year earlier, conditional on no fire being recorded at that time.
\end{itemize}
After filtering and negative data synthesis, the labeled dataset comprises $126\,800$ samples, including $50\,720$ positive (ignition) and $76\,080$ negative (non-ignition) instances.

\subsubsection{Spatiotemporal Feature Extraction}

Each pair (coordinate, datetime), positive or negative, is expanded into a $75-$day temporal window, including $60$ days prior and $15$ days after the wildfire ignites. This window captures both pre-ignition and early post-ignition meteorology.

We extracted daily data from the GRIDMET dataset~\cite{gridmet}, which offers $4 \ \rm{km}$ resolution across CONUS. The corpus contains $126\,800$ sequences, each comprised of a $75-$days window, amounting to \(9.5\) million labeled samples. Each sample is represented as a vector of meteorological features (Table~\ref{tab:dataset_snapshot}), including:

% To handle the scale of the dataset, we employed $24$ prefetch workers and $64$ parallel processing workers to efficiently stream and transform the data. Zstandard compression was used to reduce memory overhead, while caching via compressed pickle files helped minimize redundant data retrieval. All processing tasks were carried out on the CCR high-performance computing (HPC) infrastructure~\cite{ccr}.

\begin{itemize}
    \item Moisture \& Precipitation: precipitation (pr), relative humidity max/min (rmax/rmin), specific humidity (sph), vapor pressure deficit (vpd).
    \item Wind, Temperature \& Solar: wind speed (vs), air temperature max/min (tmmx/tmmn), solar radiation (srad).
    \item Fire Indices \& Evapotranspiration: burning index (bi), energy release component (erc), fuel moisture $100$hr/$1000$hr (fm$100$/fm$1000$), potential evapotranspiration (pet), actual evapotranspiration (etr).
\end{itemize}

These features enable the deep learning model to learn both short-term ignition triggers, such as sudden temperature spikes or wind shifts, and long-term stress signals such as ongoing drought conditions, wind exposure, or gradual fuel drying.

This high-resolution, spatiotemporal dataset forms the foundation for wildfire ignition prediction and realistic scenario simulation in subsequent pipeline stages.

\subsection{Wildfire Ignition Forecasting}

We designed a deep learning model to forecast wildfire ignition based on the $75-$day meteorological and environmental context window. The model treats the task as a binary classification problem: given a sequence of atmospheric and fire-related covariates, predict whether a fire will ignite at the target location (coordinate, datetime).

\begin{figure}[htbp]
    \centerline{\includegraphics[width=1\linewidth]{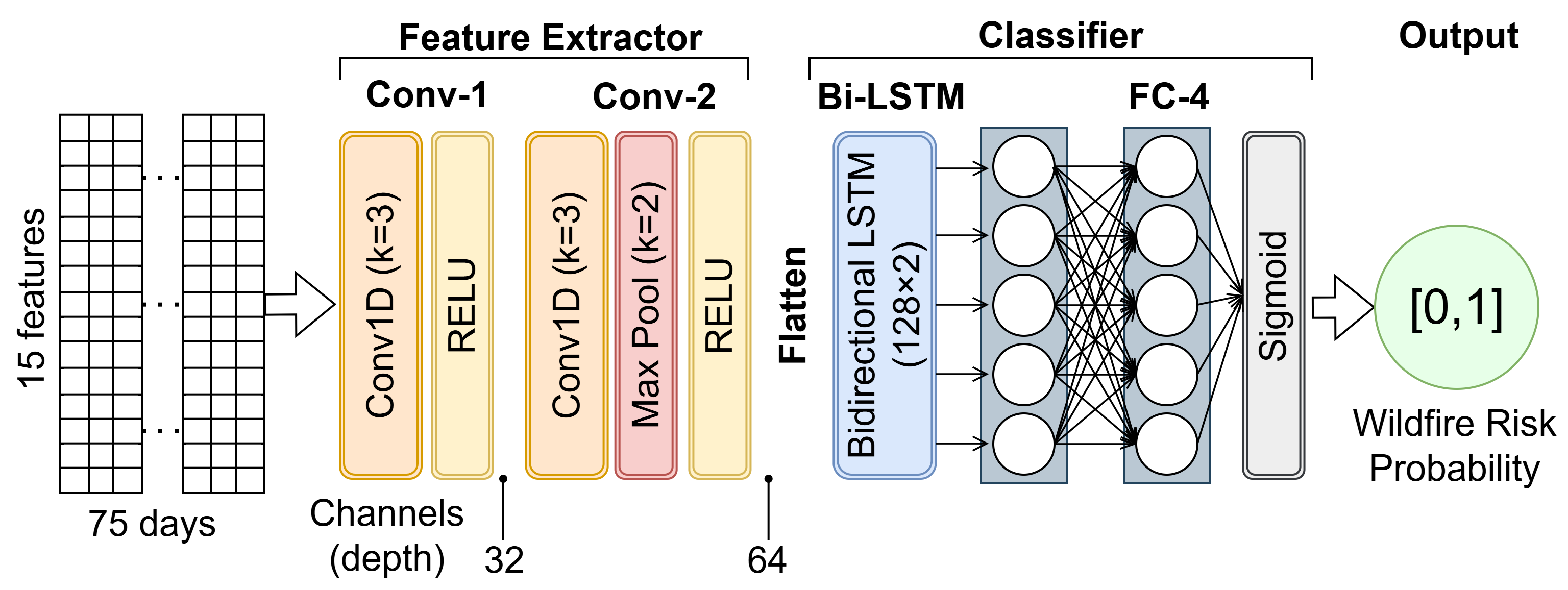}}
    \caption{The architecture of the CNN \& Bi-LSTM based neural network model for wildfire prediction.}
    % \vspace{-0.25 cm}
    \label{fig:cnn_lstm_model}
\end{figure}

We use a hybrid CNN-LSTM architecture (Fig.~\ref{fig:cnn_lstm_model}) that captures both spatial correlations and temporal dynamics to forecast a wildfire ignition. The model returns a binary output; if above a threshold ($\omega$), it is classified as ``Wildfire".

\subsection{Suppression Strategies Optimization using RL}

If the forecasted wildfire probability for a given location and time exceeds a predefined threshold ($\omega$), the framework constructs a 3D terrain model by retrieving real-world elevation, land cover, and mesoscale wind data to initiate a physics-informed fire-spread simulation. Although the simulation omits several meteorological predictors (e.g.,\ solar radiation and long-term humidity), it provides a sufficiently realistic sandbox for suppression strategy optimization.

\begin{figure*}[th]
\centerline{\includegraphics[width=1\linewidth]{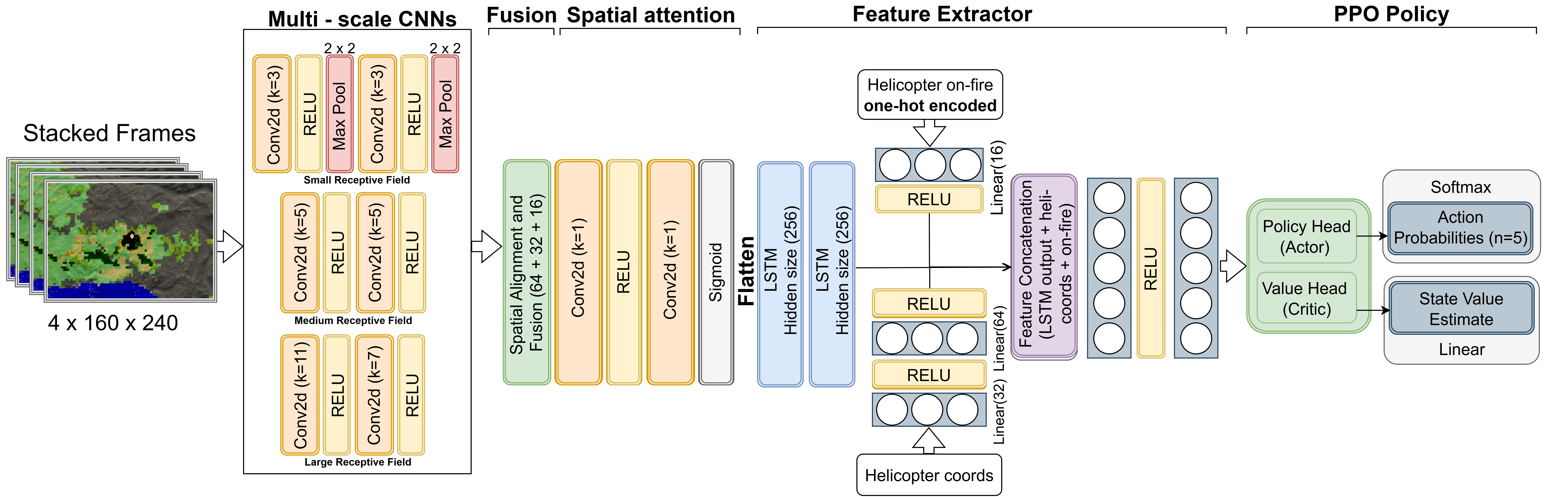}}
% \centerline{\includegraphics[width=1\linewidth]{images/main_framework_new.png}}
\caption{Architecture of the custom PPO policy network. It combines multi-scale CNNs, a spatial attention module, and LSTM layers for capturing temporal fire dynamics.}
\label{fig:ppo_flowchart}
\end{figure*}

\subsubsection{Physics-informed Simulation}

% Copy-pasted the last sentence in this paragraph from what was section b) Cellular Automata Dynamics (~line 559)
The simulation environment is based on a Cellular Automata engine, embedded in a custom Gymnasium wrapper. This setup allows the fire to evolve step-by-step while the agent interacts with the physics-informed 3D environment. Simulation state transitions evolve over discrete time steps, incorporating terrain elevation, wind direction, and land cover characteristics.

Our simulation backend is adapted from the Concord Consortium's wildfire model~\cite{concord_consortium}. The terrain is rendered as a $240$\(\times\)$160$ grid, layered with digital elevation maps and real-time land cover data.

% \paragraph{Cell Structure and State Variables}
% Each cell in the grid is modeled as a discrete CA unit with the following attributes:
% \begin{itemize}
%     \item \texttt{x}, \texttt{y}: grid coordinates
%     \item \texttt{zone}: terrain and vegetation type with fuel properties
%     \item \texttt{baseElevation}: ground elevation (m)
%     \item \texttt{spreadRate}: local fire spread velocity (m/s)
%     \item \texttt{burnTime}: time remaining until burnout
%     \item \texttt{fireState}: one of \texttt{Unburnt}, \texttt{Burning}, or \texttt{Burnt}
%     \item \texttt{ignitionTime}: simulation timestamp of ignition
%     \item \texttt{helitackDropCount}: suppressant drop count
%     \item \texttt{isFireSurvivor}: boolean flag for fire escape
% \end{itemize}

% Simulation state transitions evolve over discrete time steps, incorporating terrain slope, wind direction, and vegetation characteristics.

\paragraph{Fire Engine and Spread Modeling}
We use a fire engine module based on Rothermel’s fire spread equations~\cite{rothermel_fire_spread_eq}, adapted for standard grid-based wildfire simulators such as FARSITE~\cite{finney2004farsite}. The base rate of spread ($R$) is computed as:
\begin{equation}
R = \frac{I_R \cdot \xi}{\rho_b \cdot \epsilon \cdot Q_{ig}}
\end{equation}
where $I_R$ is reaction intensity, $\xi$ is the propagating flux ratio, $\rho_b$ is bulk density, $\epsilon$ is the heating number, and $Q_{ig}$ is the ignition heat.

To capture how wind and terrain accelerate fire progression, the model adjusts the base rate of spread ($R$) as follows:
\begin{equation}
R_{\text{eff}} = R (1 + \Phi_w + \Phi_s)
\end{equation}
where $\Phi_w$, $\Phi_s$ are wind and elevation adjustments.

\paragraph{Fuel Models and Parameters}
% Each vegetation type includes the following properties:
% \begin{itemize}
%     \item \texttt{SAV}: Surface area-to-volume ratio
%     \item \texttt{Net Fuel Load}: Dry fuel available
%     \item \texttt{Fuel Bed Depth}: Vertical extent of combustibles
%     \item \texttt{Packing Ratio}: Fuel density
%     \item \texttt{Moisture of Extinction(mx)}: Minimum moisture for combustion
% \end{itemize}
% These values are derived from Anderson’s $13$ Fuel Models and Scott \& Burgan’s 40 Fuel Models and normalized for our simulation.

% Updated to be shorter in a sentence format.

Each land cover represents a vegetation type that includes
surface area-to-volume ratio, dry fuel availability, vertical extent of combustibles, fuel density, and minimum moisture for combustion. These values are derived from Anderson’s $13$ Fuel Models~\cite{anderson1982aids} and Scott \& Burgan’s $40$ Fuel Models~\cite{scott2005standard}, and adjusted for our simulation.

\paragraph{Realistic Terrain and Vegetation Integration}
% Replaced "See Table II" with (Table II)

To generate a realistic simulation environment, we construct a 3D terrain for each high-risk forecast (visualized in Table~\ref{tab:wildfire_environment_creation}). Using the Google Earth Engine~\cite{earth_engine}, we source MODIS land cover, SRTM elevation data, and GRIDMET~\cite{gridmet} wind data.

\begin{table*}[htbp]
\centering
\caption{
High-fidelity simulation environment creation stages: (1) region selection, (2) extracted land cover raster, (3) corresponding elevation map, and (4) the 3D rendered simulation environment with terrain, land cover, and elevations.}
\begin{tabular}{cccc}
\toprule
\textbf{Region} & \textbf{Landcover} & \textbf{Elevation map} & \textbf{3D Render} \\
\midrule
\begin{minipage}[t]{0.19\textwidth}
    \centering
    \includegraphics[width=\linewidth]{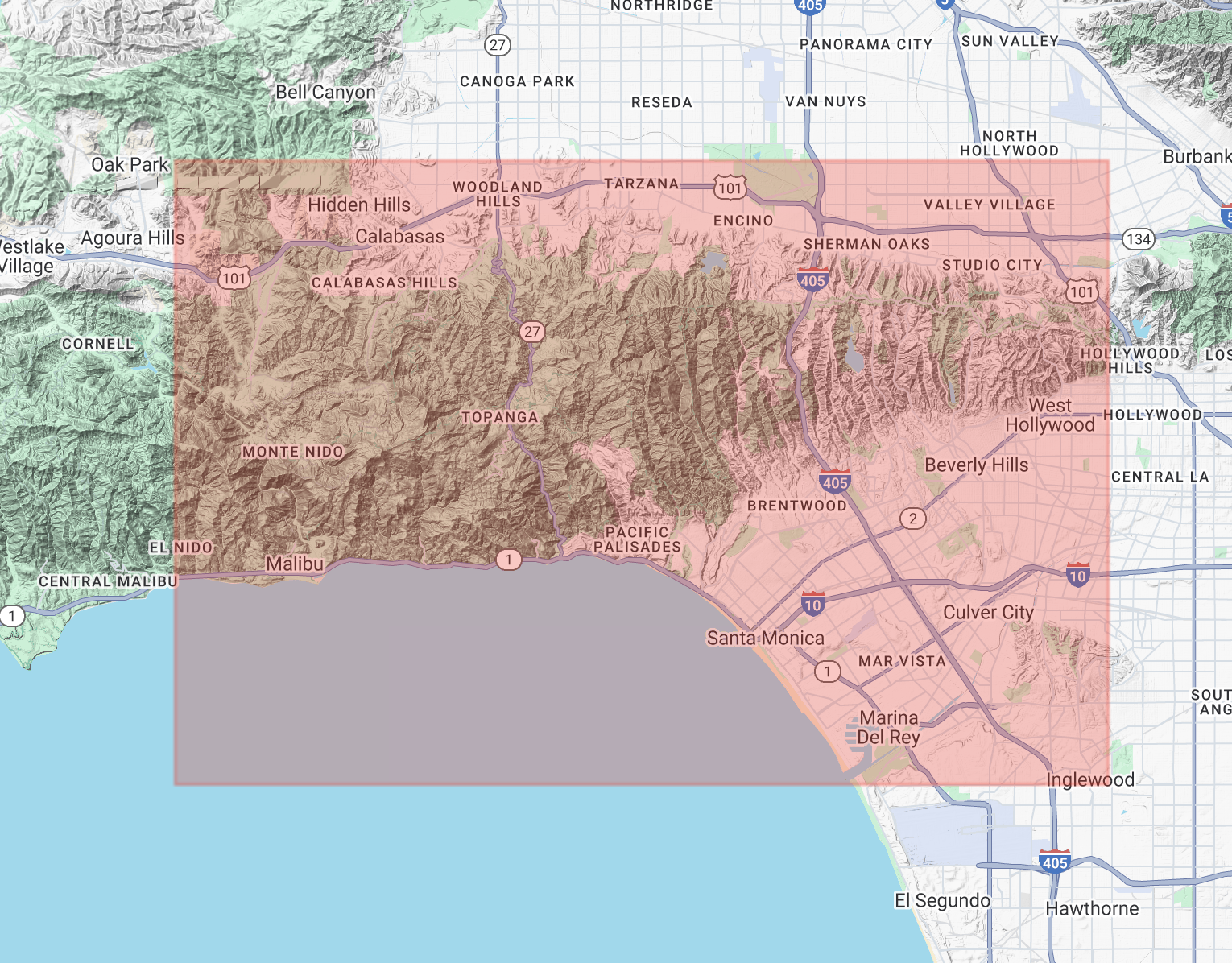}\\
    %\vspace{2pt}
    {\footnotesize Region in Los Angeles, CA}
\end{minipage} &
\includegraphics[width=0.22\textwidth]{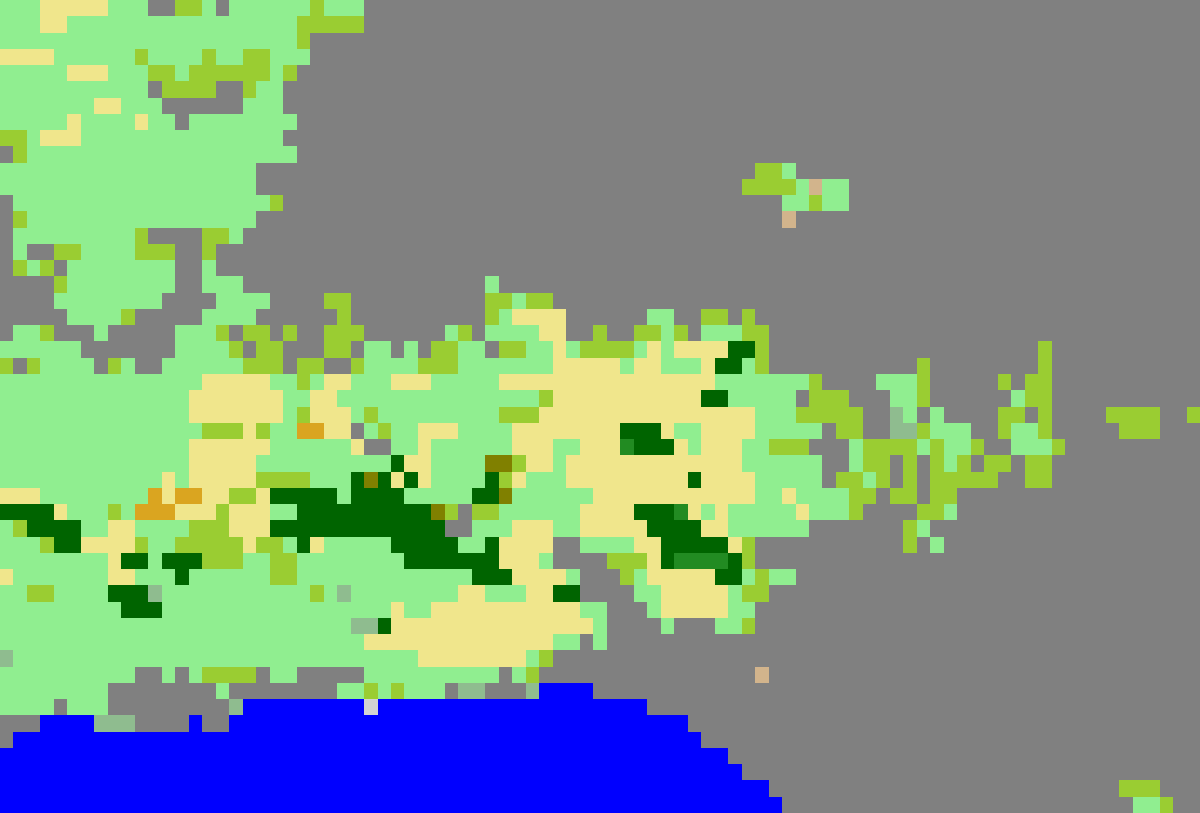} & 
\includegraphics[width=0.22\textwidth]{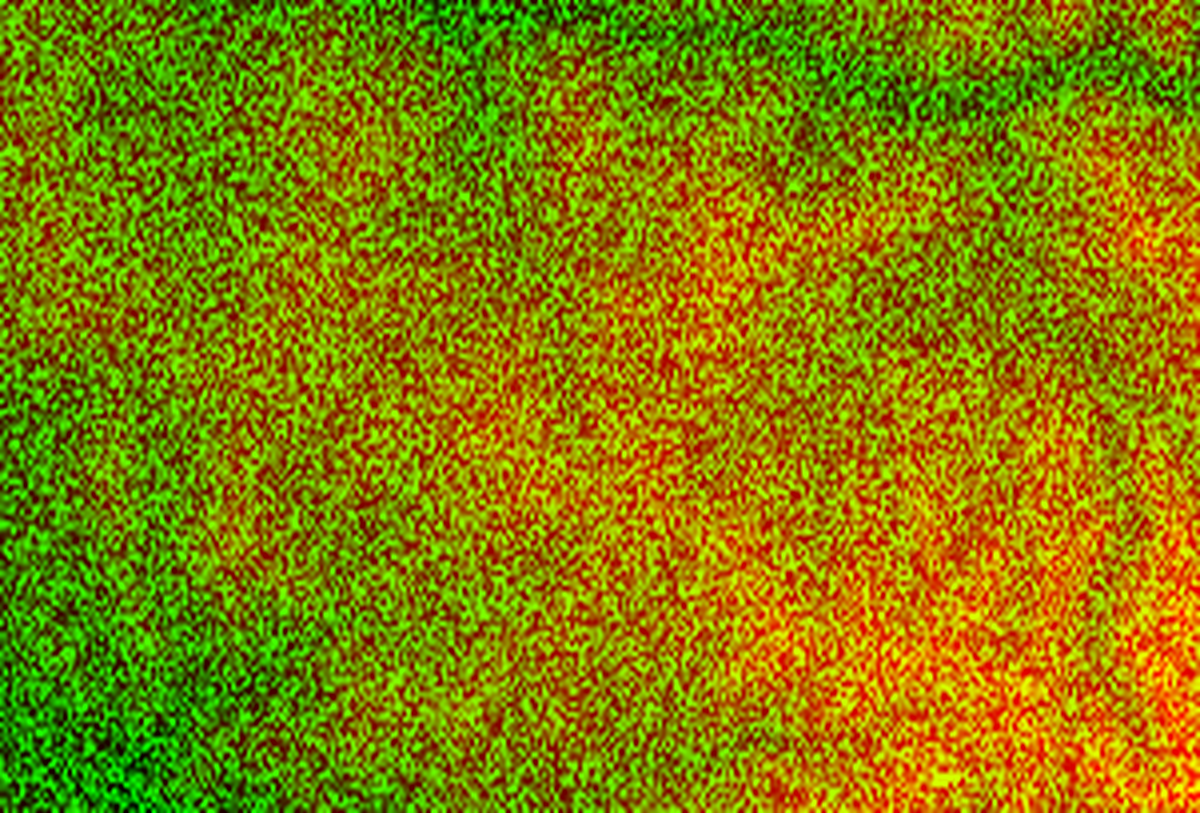} & 
\includegraphics[width=0.25\textwidth]{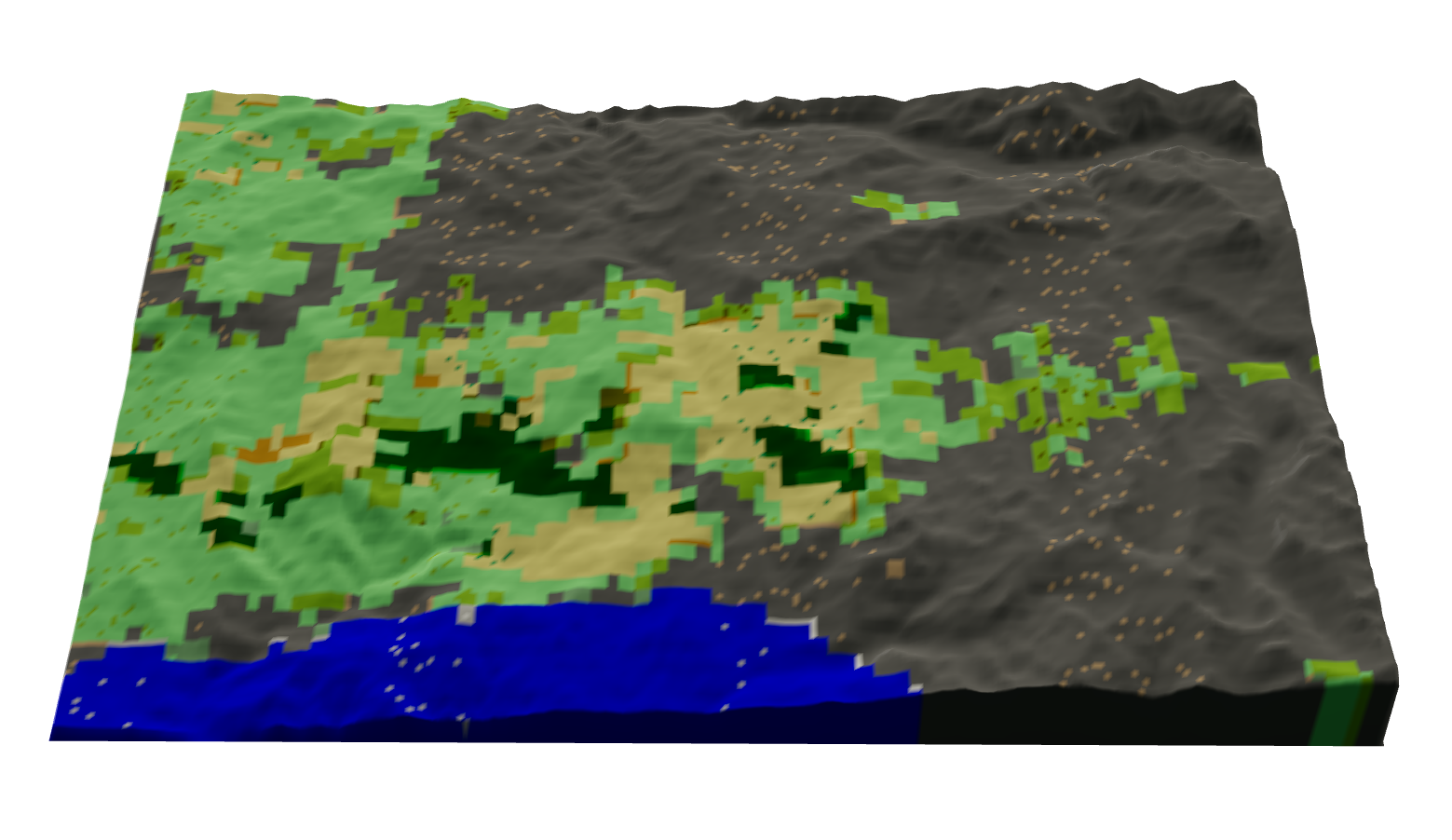} \\
\midrule
\begin{minipage}[t]{0.20\textwidth}
    \centering
    \includegraphics[width=\linewidth]{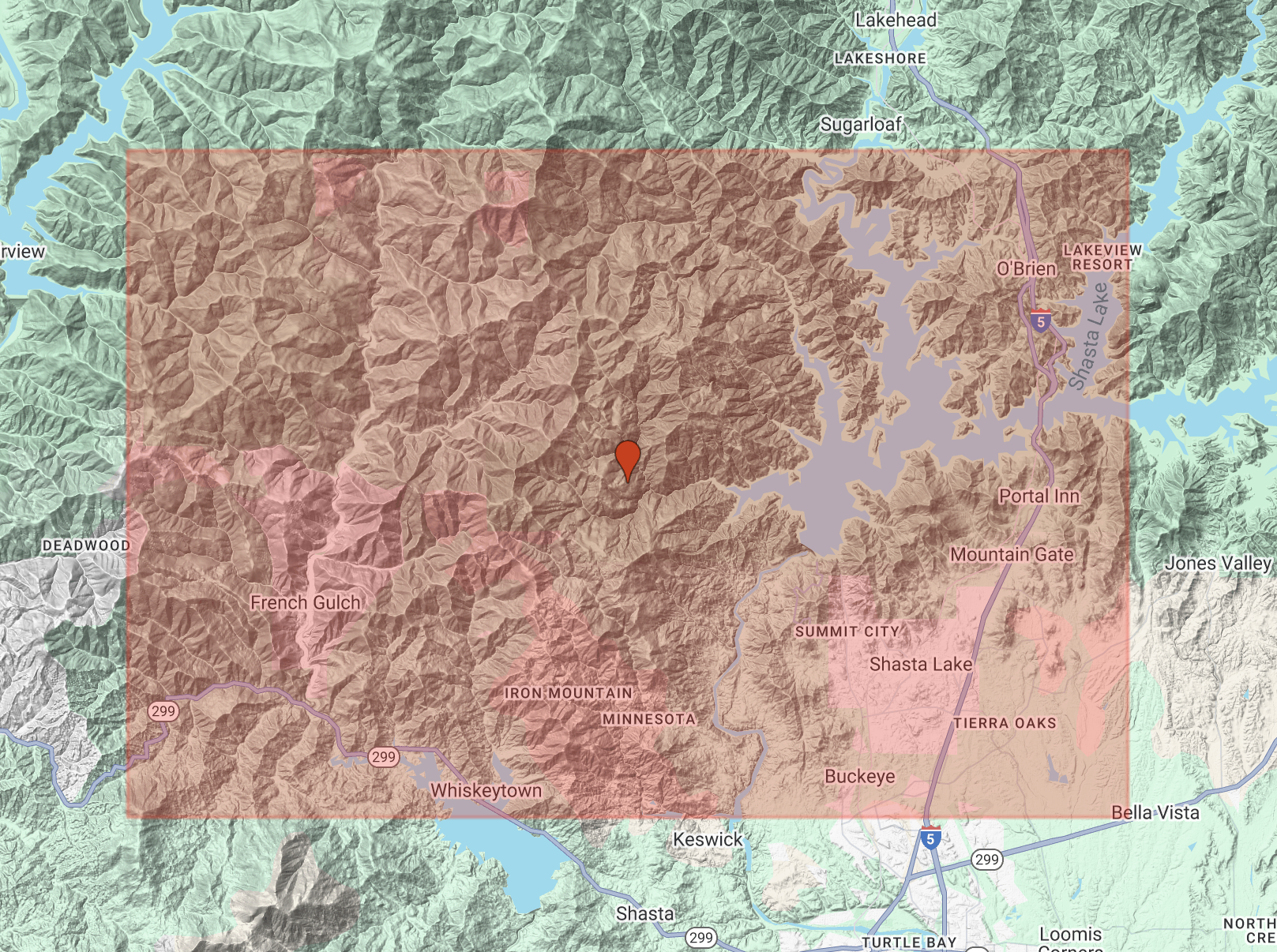}\\
    % \vspace{2pt}
    {\footnotesize Region in French Gulch, CA}
\end{minipage} &
\includegraphics[width=0.22\textwidth]{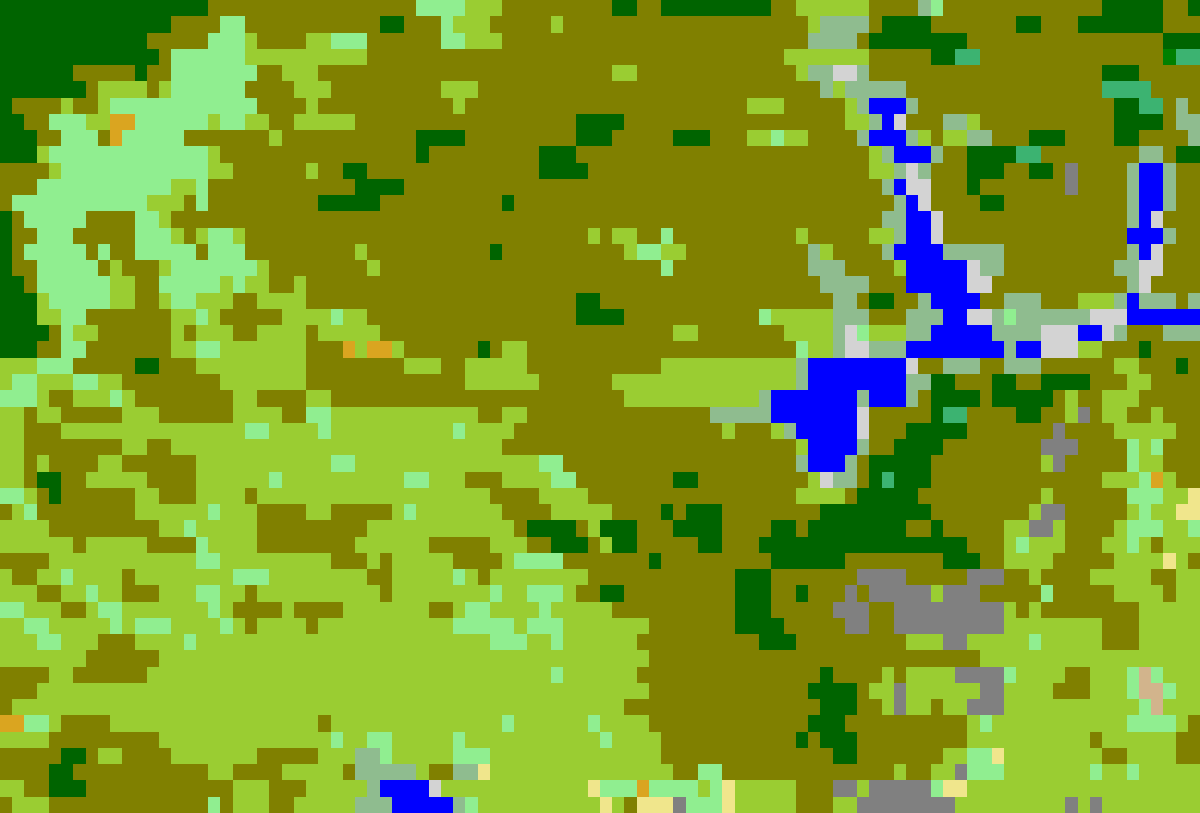} & 
\includegraphics[width=0.22\textwidth]{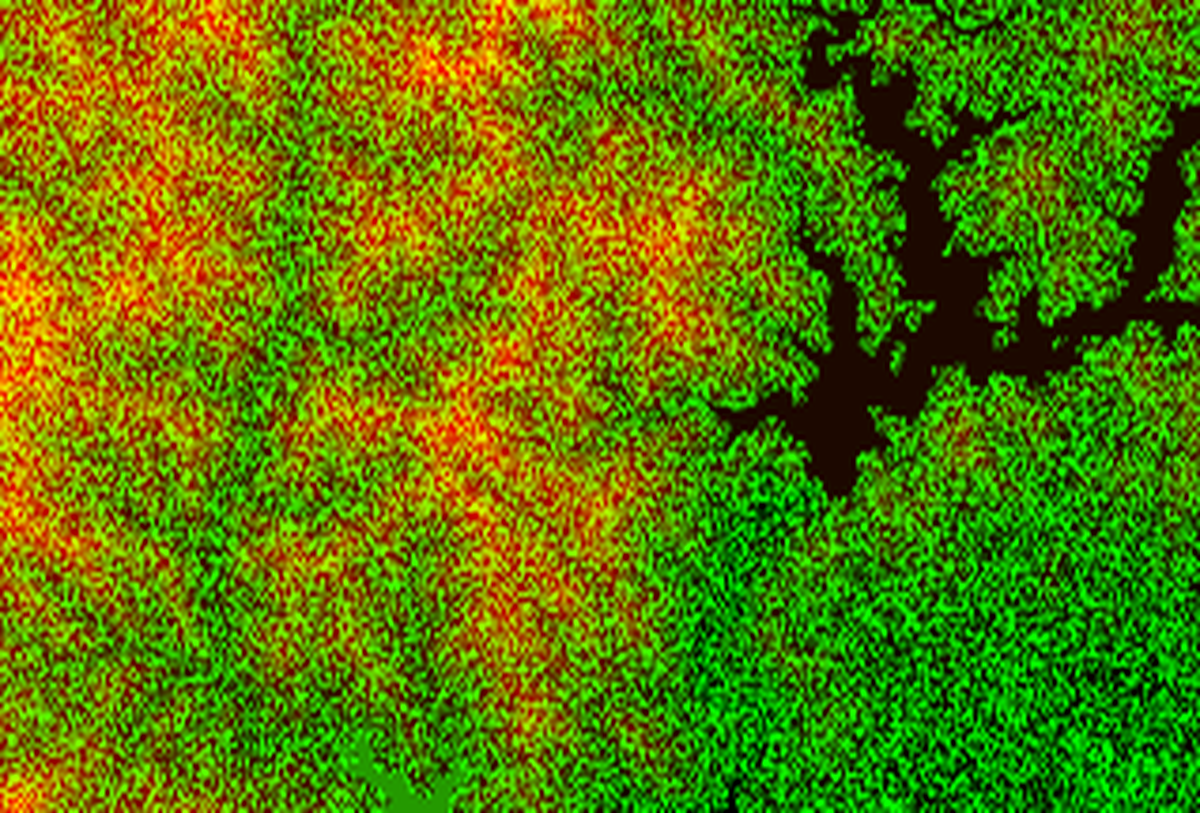} & 
\includegraphics[width=0.25\textwidth]{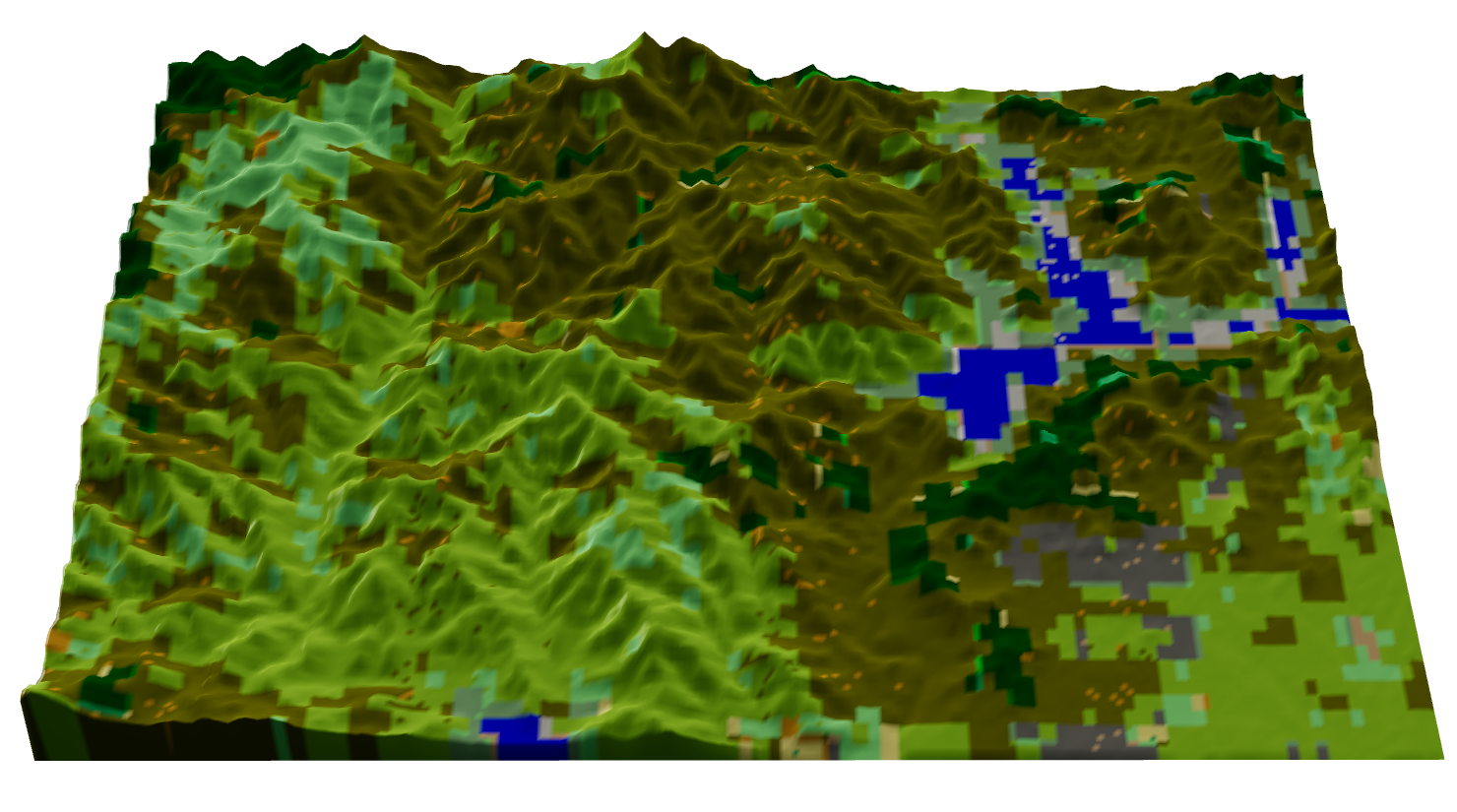} \\
\bottomrule
\end{tabular}

\label{tab:wildfire_environment_creation}
\end{table*}

\subsubsection{Custom RL Environment Setup}

The agent represents a single helitack unit, an aerial firefighting resource tasked with navigating the environment and deploying fire suppressant.
\paragraph{Action Space} The environment has a discrete action space: (\texttt{Up}, \texttt{Down}, \texttt{Left}, \texttt{Right}) and one for releasing suppressant (\texttt{Drop}).

\paragraph{Observation Space}
The observation space includes:
\begin{itemize}[leftmargin=1.2cm]
    \item A $4$-frame stack of $160$\(\times\)$240$ environment grids capturing fire states (unburnt, burning, and burnt) and intensities over time.
    \item Current agent position in the grid.
    \item Binary flag indicating if the agent is positioned over a burning cell.
\end{itemize}

\paragraph{Reward Function}
The reward function is based on short-term suppression with long-term containment by considering both state-based and proximity-aware feedback.
\begin{itemize}[leftmargin=1.2cm]
    \item Positive rewards: extinguish burning cells; prevent fire expansion; maintain proximity to fire fronts.
    \item Negative rewards: fire growth or large burnt area; inaction or delayed intervention; hovering over burnt or inert terrain.
\end{itemize}

This encourages the intelligent suppression strategies such as dynamic circling of fire fronts, firebreak creation, and early response to high-risk areas.

\paragraph{Feature Extractor} We extract a condensed feature vector from the structured observations using a custom multi-stage feature extractor. First, the stacked environment grids are processed by three parallel CNN branches with different receptive fields to capture fire patterns at multiple spatial scales. The resulting feature maps are unified and then refined by a spatial attention module (Eq.~\ref{eq:attention_map}) to focus the network on the most critical fire zones. Finally, the flattened spatial features are fed into a two-layer LSTM, which integrates temporal information by learning from the historical dynamics of the fire and the agent's past actions.
\begin{equation}
\label{eq:attention_map}
\text{AttendedMap} = \sigma\left( \text{Conv}_{1 \times 1} \left( \text{ReLU} \left( \text{Conv}_{1 \times 1}(F) \right) \right) \right)
\end{equation}

Additionally, we process helitack coordinates passed through a multi-layer perceptron (MLP) encoder and a binary flag indicating if our RL agent is over a burning grid cell.

Finally, we concatenate all extracted features into a unified vector. This fused representation is then passed through a $2-$layer fully connected residual block to produce the latent feature vector that serves as the input to the RL policy and value networks. A detailed architecture is presented in Fig.~\ref{fig:ppo_flowchart}.

\subsubsection{RL Algorithm}

We use the Proximal Policy Optimization (PPO) \cite{schulman2017ppo} algorithm to train our agent. PPO's stable learning process is well-suited for our dynamic wildfire suppression environment because it effectively handles complex state transitions influenced by agent actions and sparse rewards.

\subsection{Fire Threat Assessment Report}
During each validation rollout, we log: (1) coordinates and timestamps of each suppressant drop; (2) burnt area trajectory; (3) predicted ignition coordinates, time of ignition, and confidence score from the forecasting stage; (4) number of helitack deployments, time elapsed before containment; and (5) suggested evacuation advisories, suppression prioritization zones, and contingency thresholds.

We use these metrics to prepare a fire threat assessment report that offers decision-makers a real-time operational summary. It includes forecasted ignition patterns with the simulated suppression plan.

\subsection{Web-based Application}
We make our framework accessible to decision-makers, researchers, and the public by developing an interactive web application that serves as a front-end to the entire \textit{FireCastRL} pipeline.

\section{Results}

% We evaluate the performance of our proposed \textit{FireCastRL} framework across three core stages.

% : (A) data acquisition and preparation, (B) wildfire ignition forecasting (C) supression strategies optimization using RL.

\subsection{Dataset Release for Community Research}
To foster community research, we are releasing a public spatiotemporal dataset containing $9.5$ million labeled records, structured as $75-$day multivariate time series windows that capture both wildfire ignition and non-ignition events.

Each sample includes $15$ environmental features (e.g., precipitation, wind speed, burn index, humidity) extracted from GRIDMET. The dataset is publicly available at \href{https://www.kaggle.com/datasets/firecastrl/us-wildfire-dataset}{https://www.kaggle.com/datasets/firecastrl/us-wildfire-dataset}.

% The dataset (https://www.kaggle.com/datasets/firecastrl/us-wildfire-dataset), covering the contiguous United States is published on Kaggle and includes pre-processed data, normalization stats, and metadata.

\subsection{Wildfire-Ignition Forecasting Results}
We trained a hybrid CNN-LSTM classifier on $75-$day multivariate time series window. The model encodes spatial structure via CNNs and temporal evolution via LSTMs. We applied focal loss and class-balanced sampling to address ignition imbalance. Notably, the model predicted the Palisades wildfire (January 2025) with $98.6\%$ confidence. 

The confusion matrix contained $536$ true positives, $1251$ false negatives, $1455$ false positives, and $458$ true negatives. The model achieved $73.1\%$ accuracy on a held-out test set (January 2025 to April 2025) (Table~\ref{tab:cls_metrics}). The model's performance was primarily impacted by unpredictable, human-caused ignitions that are independent of weather patterns. This highlights an important direction for future work: incorporating data on human activity to achieve more comprehensive prediction.

\begin{table}[htbp]
\caption{Wildfire ignition model comparison}
\centering
\begin{tabular}{l c c c c}
\toprule
\textbf{Model} & \textbf{Accuracy [\%]} & \textbf{Precision} & \textbf{Recall} & \textbf{F$_1$} \\
\midrule
\rowcolor{gray!20}
\textbf{CNN–LSTM (ours)} & \textbf{73.1} & \textbf{0.71} & \textbf{0.70} & \textbf{0.70} \\
XGBoost* & 66.4 & 0.65 & 0.56 & 0.60 \\
Gradient Boosting* & 65.6 & 0.66 & 0.49 & 0.56 \\
Random Forest* & 64.6 & 0.61 & 0.59 & 0.60 \\
K-Nearest Neighbors* & 63.7 & 0.60 & 0.57 & 0.59 \\
Simple-MLP* & 62.6 & 0.57 & 0.65 & 0.61 \\
Decision Tree* & 62.0 & 0.58 & 0.58 & 0.58 \\
Two-Layer-LSTM & 61.9 & 0.62 & 0.55 & 0.57 \\
LightTS-Inspired & 60.6 & 0.54 & 0.79 & 0.64 \\
Logistic Regression* & 59.7 & 0.55 & 0.62 & 0.58 \\
Naive Bayes* & 56.5 & 0.52 & 0.58 & 0.54 \\
\bottomrule
\end{tabular}
% \vspace{-6pt}
\label{tab:cls_metrics}
\caption*{\footnotesize * Baselines not inherently time-series-aware; feature vectors flattened to $1125$ dimensions ($15$ variables \(\times\) $75$ timesteps).}
% \vspace{-0.35 cm}
\end{table}

\begin{table*}[htbp]
\centering
\caption{Comparison between trained PPO agent and rule-based agent}
\label{tab:agent_comparison}
\renewcommand{\arraystretch}{1.2}
\begin{tabular}{
    >{\centering\arraybackslash}m{4.2cm}
    >{\raggedright\arraybackslash}m{3.8cm}
    >{\centering\arraybackslash}m{4.2cm}
    >{\raggedright\arraybackslash}m{3.8cm}
}
\toprule
\multicolumn{2}{c}{\textbf{Trained PPO Agent}} & \multicolumn{2}{c}{\textbf{Rule-Based Agent}} \\
\midrule
\includegraphics[width=1.05\linewidth]{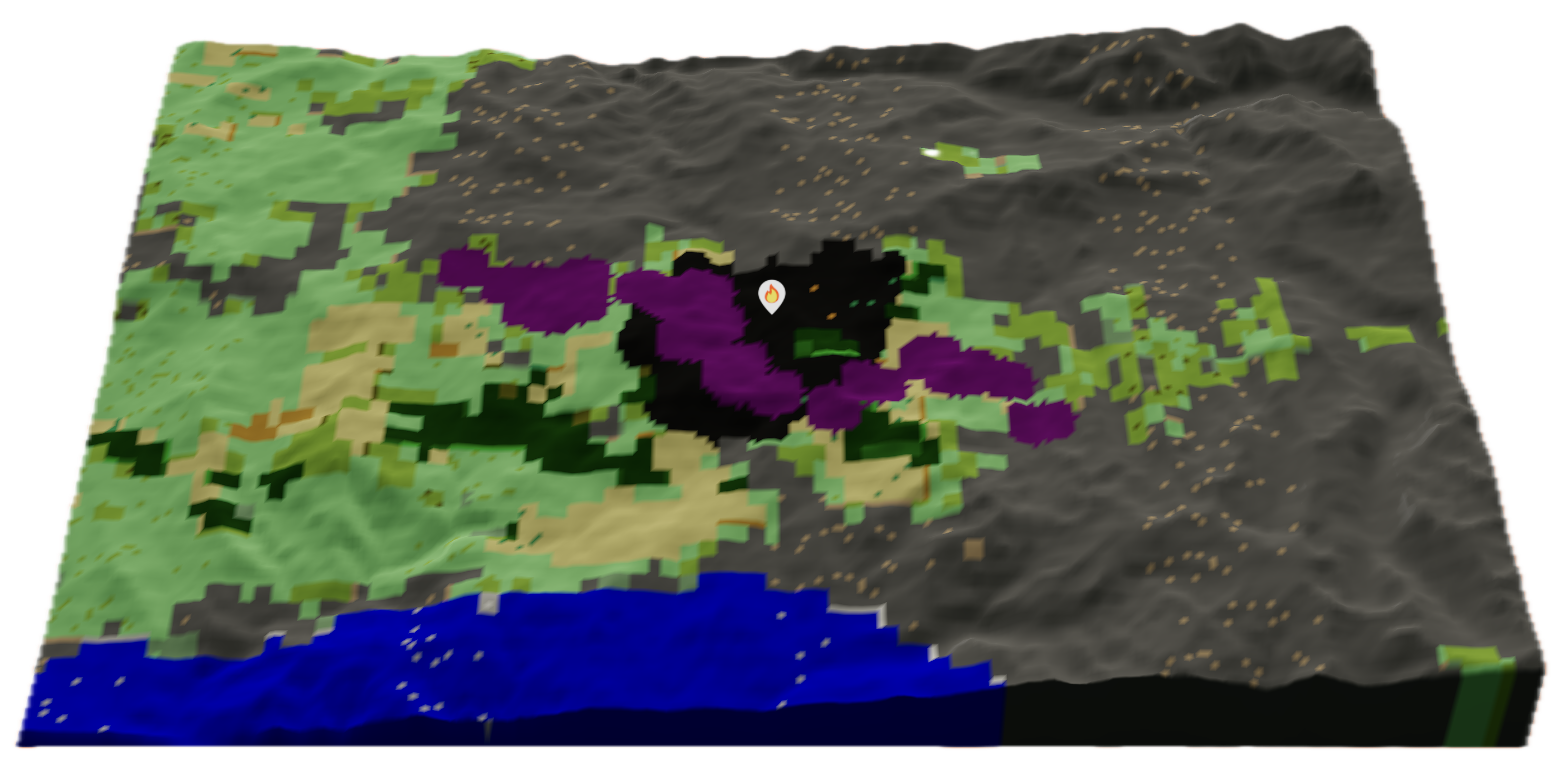} & 
\begin{tabular}{@{}ll}
    \textbf{Cells Burned:} & 1529 \\
    \textbf{Timesteps:} & 410 \\
    \textbf{Helitacks:} & 18 \\
    \textbf{Water Used:} & $14\,400$ gal \\
\end{tabular} &
\includegraphics[width=1.05\linewidth]{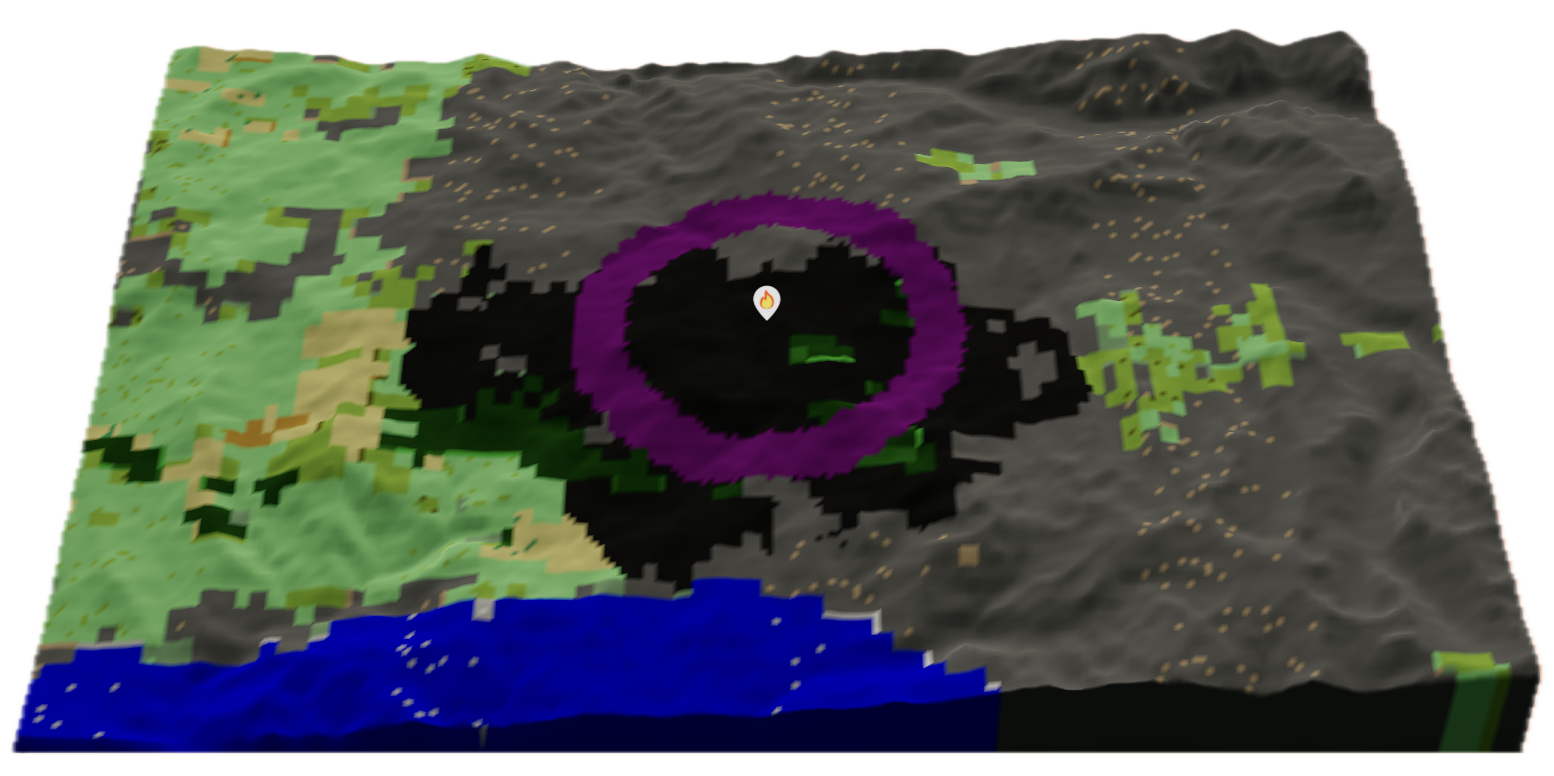} & 
\begin{tabular}{@{}ll}
    \textbf{Cells Burned:} & 4931 \\
    \textbf{Timesteps:} & 883 \\
    \textbf{Helitacks:} & 47 \\
    \textbf{Water Used:} & $37\,600$ gal \\
\end{tabular} \\
\bottomrule
\end{tabular}
\captionsetup{justification=centering}
\caption*{\footnotesize\textit{Note:} Purple cells represent areas where helitack agents deployed suppressant.}
% \vspace{-0.38cm}
\end{table*}

\subsection{Helitack Suppression Performance}

% \subsubsection{Training Performance} After $3{\times}10^{5}$ training steps, our PPO-based helitack agent reliably learned a circling behavior around the flame front, effectively slowing fire progression in held-out simulation maps. Table ~\ref{tab:agent_comparison} present the agent comparison results.

% \subsubsection{Helitack Agent Evaluation}

% We compared the trained agent to a rule-based one that circles the fire and drops suppressant without using any observation of the fire's state (Table~\ref{tab:agent_comparison}).

% These results show that combining a 73.1\% accurate forecaster with an RL agent enables effective suppression strategies and reduces projected damage, while also offering a valuable benchmark dataset for the community.

Hyperparameters for training the PPO algorithm:
{ 
$n_{\mathrm{steps}}=128$, batch size $=64$, $n_{\mathrm{epochs}}=3$, learning rate $=3\times10^{-4}$, clip range $=0.1$, $\gamma=0.95$, GAE--$\lambda=0.9$, entropy coefficient $=0.2$, value-function coefficient $=0.4$, max-grad-norm $=1.0$, and target-KL $=0.03$.
}

After $3 \times 10^{5}$ training steps, our PPO-based helitack agent reliably learned to circle the flame front, effectively slowing fire progression on held-out simulation maps. We compared this agent to a rule-based baseline that drops suppressant without observing the fire's state, with results presented in Table~\ref{tab:agent_comparison}. These results show that combining an accurate forecaster with an RL agent enables effective suppression strategies, reduces projected damage, and provides a valuable benchmark dataset for the community.

% \begin{table*}[htbp]
% \centering
% \caption{Comparison between Trained PPO Agent and Rule-Based Agent.}
% \renewcommand{\arraystretch}{1.1}
% \begin{tabular}{
%     >{\arraybackslash}m{4cm} 
%     >{\centering\arraybackslash}m{5.5cm} 
%     >{\centering\arraybackslash}m{5.5cm}
% }
% \toprule
% \textbf{Metric} & \textbf{Trained PPO Agent} & \textbf{Rule-Based Agent} \\
% \midrule
% Simulation Screenshot & 
% \includegraphics[width=0.2\textwidth]{images/trained_agent_sim.png} & 
% \includegraphics[width=0.2\textwidth]{images/rule_based_agent_sim.png} \\
% \midrule
% Cells Burned in Episode & 1529 & 4931 \\
% \midrule
% Timesteps & 410 & 883 \\
% \midrule
% Number of Helitacks & 18 & 47\\
% \midrule
% Total Water Used (gallons) & $14\,400$ & $37\,600$\\
% % \midrule
% % Water Efficiency (gallons per cells saved) & 0.39 &  1.12\\
% \bottomrule
% \end{tabular}
% \label{tab:agent_comparison}
% % \captionsetup{justification=centering}
% \caption*{\footnotesize\textit{Note:} Purple cells represent areas where helitack agents deployed suppressant.}
% \vspace{-0.25 cm}
% \end{table*}

\subsection{FireCastRL Web Application}

We provide a web interface to demonstrate the complete \textit{~FireCastRL} pipeline. Users can select a location to generate an ignition risk forecast and, for high-risk scenarios, launch a simulation to visualize the RL agent's optimal suppression response. It also generates the final threat assessment report. Details are available at \href{https://sites.google.com/view/firecastrl}{https://sites.google.com/view/firecastrl}.

% \vspace{-0.4cm}
\section{Conclusion and Future Work}
% We introduced FireCastRL, an integrated pipeline for proactive wildfire management that combines a deep learning forecasting model with an RL suppression agent in a realistic, data-driven simulation. Our system predicts ignition hotspots using large-scale historical data and then trains an agent to execute effective suppression strategies in environments built from real-world terrain and weather. This framework can help develop novel tactics and assist decision-makers in optimizing resource allocation via generated Threat Assessment Reports.

% Future work will focus on two key areas: enhancing the forecasting model with advanced architectures like Transformers and GANs, and expanding the simulation to include diverse assets such as ground crews and firebreaks, requiring the agent to learn complex coordination strategies.

In this paper, we presented \textit{FireCastRL}, a pipeline that combines wildfire prediction and intelligent suppression in a realistic simulation setting. We trained a deep learning model using over 9.5 million real-world data samples to forecast where wildfires are likely to start, and followed that up with an RL agent that learns how to fight fires in those predicted areas. The simulation uses real terrain and weather data to give the agent a realistic environment to train in, and the whole system works together to move wildfire response from reactive to proactive.

Our RL simulation can be used to develop effective suppression strategies in varied terrains and conditions. The threat assessment report can assist firefighters and policy-makers in planning a response by prioritizing high-risk zones and optimizing resource allocation.

Future work can focus on improving the forecasting model with advanced architectures like Transformers, accounting for human-caused ignitions, extending the simulation to a multi-agent setup, and including diverse suppression strategies such as ground crews and firebreaks.

% , requiring the agent to learn complex coordination strategies.

% \vspace{-0.4em}
\bibliographystyle{IEEEtran}
\bibliography{references}

\end{document}